\documentclass[10pt,twocolumn,letterpaper]{article}

\usepackage{cvpr}
\usepackage{times}
\usepackage{epsfig}
\usepackage{graphicx}
\usepackage{amsmath}
\usepackage{amssymb}
\usepackage{multirow}

\usepackage[colorlinks,linkcolor=black,anchorcolor=black,citecolor=green]{hyperref}

\cvprfinalcopy 
\def\etal{\emph{et al.}}
\def\eg{\emph{e.g.}}  \def\ie{\emph{i.e.}}
\def\etc{\emph{etc.}} \def\vs{\emph{vs}.}

\def\cvprPaperID{1156} 

\ifcvprfinal\fi
\begin{document}

\title{Scale-recurrent Network for Deep Image Deblurring}

\author{Xin Tao$^{1,2}$ \quad Hongyun Gao$^{1,2}$ \quad Yi Wang$^1$ \quad Xiaoyong Shen$^2$ 
\quad Jue Wang$^3$ \quad Jiaya Jia$^{1,2}$\\
$^1$The Chinese University of Hong Kong\quad$^2$Youtu Lab, Tencent\quad$^3$Megvii Inc.\\
}

\maketitle

\begin{abstract}
  In single image deblurring, the ``coarse-to-fine'' scheme, \ie~gradually restoring the sharp image on different resolutions in a pyramid, is very successful in both traditional optimization-based methods and recent neural-network-based approaches. In this paper, we investigate this strategy and propose a Scale-recurrent Network (SRN-DeblurNet) for this deblurring task. 
  Compared with the many recent learning-based approaches in \cite{nah2017deep}, it has a simpler network structure, a smaller number of parameters and is easier to train.
  We evaluate our method on large-scale deblurring datasets with complex motion. Results show that our method can produce better quality results than state-of-the-arts, both quantitatively and qualitatively.
\end{abstract}

\section{Introduction} \label{sec:intro}
Image deblurring has long been an important problem in computer vision and image
processing. Given a motion- or focal-blurred input image, caused by camera shake, object motion or out-of-focus, the goal of deblurring is to recover a sharp latent image with necessary edge structures and details.

Single image deblurring is highly ill-posed. Traditional methods apply various constraints to model characteristics of blur (\eg~uniform/non-uniform/depth-aware), and utilize different natural image priors \cite{bahat2017non,chan1998total,cho2009fast,goldstein2012blur,pan2014deblurring,xu2010two,xu2013unnatural}
to regularize the solution space. Most of these methods involve intensive, sometimes heuristic, parameter-tuning and expensive computation. Further, the simplified assumptions on the blur model often hinder their performance on real-word examples, where blur is far more complex than modeled and is entangled with in-camera image processing pipeline.

Learning-based methods have also been proposed for deblurring. Early methods \cite{schuler2016learning,sun2015learning,xiao2016learning} substitute a few modules or steps in traditional frameworks with learned parameters to make use of external data. More recent work started to use end-to-end trainable networks for image \cite{nah2017deep} and video \cite{hyun2017online,su2017deep} deblurring. Among them, Nah \etal \cite{nah2017deep} have achieved state-of-the-art results using a multi-scale convolutional neural network (CNN). Their method commences from a very coarse scale of the blurry image, and progressively recovers the latent image at higher resolutions until the full resolution is reached. This framework follows the multi-scale mechanism in traditional methods, where the coarse-to-fine pipelines are common when handling large blur kernels~\cite{cho2009fast}.


In this paper, we explore a more effective network structure for multi-scale image deblurring. We propose the new scale-recurrent network (SRN), which discusses and addresses two important and general issues in CNN-based deblurring systems.

\begin{figure}[t]
  \begin{center}
    \includegraphics[width=1.0\linewidth]{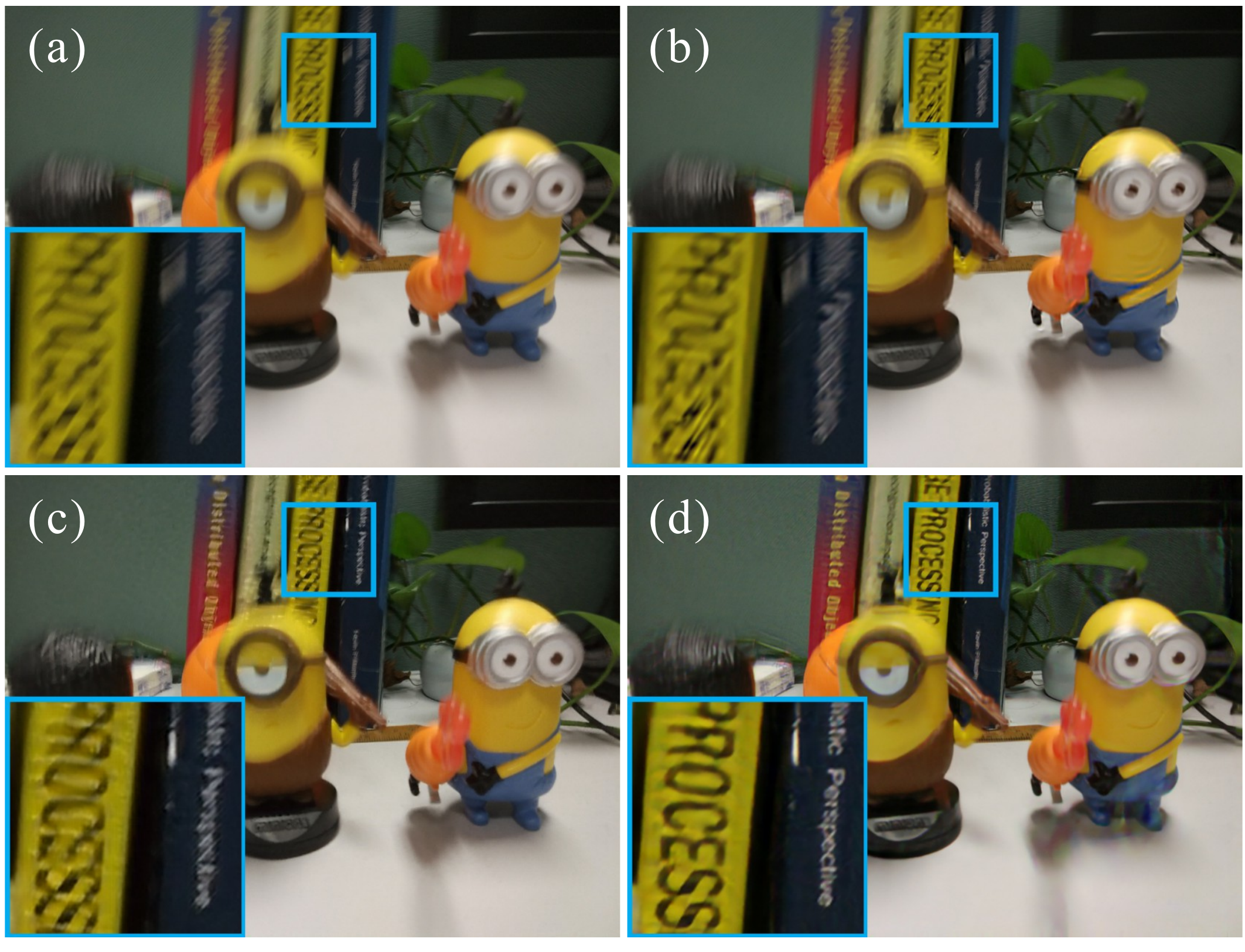}
  \end{center}
  \vspace{-0.1in}
  \caption{\textbf{One real example.} (a) Input blurred image. (b) Result of Sun \etal~\cite{sun2015learning}.
  (c) Result of Nah \etal~\cite{nah2017deep}. (d) Our result.}\label{fig:teaser}
\end{figure}

\begin{figure*}[ht]
    \begin{center}
      \includegraphics[width=1.0\linewidth]{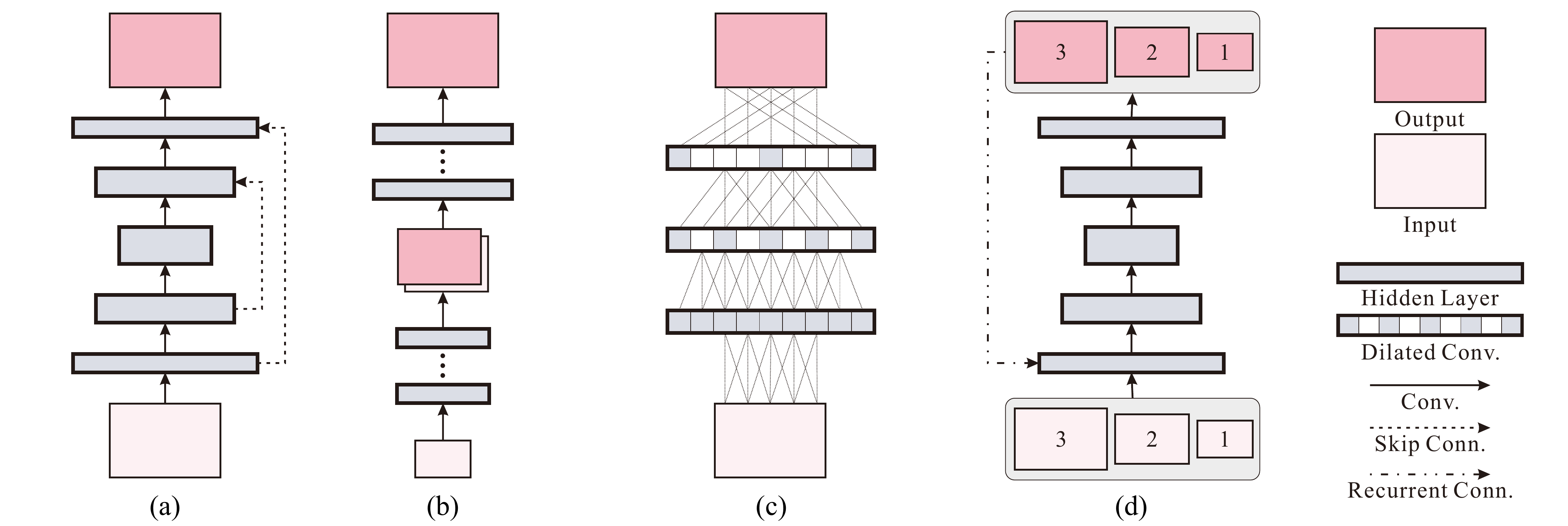}
    \end{center}
   \caption{\textbf{Different CNNs for image processing.} (a) U-Net \cite{ronneberger2015u} or encoder-decoder network \cite{mao2016image}. (b) Multi-scale \cite{nah2017deep} or cascaded refinement network \cite{chen2017photographic}. (c) Dilated convolutional network \cite{chen2017fast}. (d) Our proposed scale-recurrent network (SRN).}\label{fig:cnns}
\end{figure*}

\vspace{-0.1in}
\paragraph{Scale-recurrent Structure}
In well-established multi-scale methods, the solver and its parameters at each scale are usually the same. This is intuitively a natural choice since in each scale we aim to solve the same problem. It was also found that varying parameters at each scale could introduce instability and cause the extra problems of unrestrictive solution space. Another concern is that input images may have different resolutions and motion scales. If parameter tweaking in each scale is allowed, the solution may overfit to a specific image resolution or motion scale.

We believe this scheme should also be applied to CNN-based methods for the same reasons. However, recent cascaded networks \cite{chen2017photographic,nah2017deep} still use independent parameters for each of their scales. 
In this work, we propose sharing network weights across scales to significantly reduce training difficulty and introduce obvious stability benefits.

The advantages are twofold. First, it reduces the number of trainable parameters significantly. 
Even with the same training data, the recurrent exploitation of shared weights works in a way similar to using data multiple times to learn parameters, which actually amounts to data augmentation regarding scales. Second, our proposed structure can incorporate recurrent modules, the hidden state of which implicitly captures useful information and benefits restoration across scales.

\vspace{-0.1in}
\paragraph{Encoder-decoder ResBlock Network}
Also inspired by recent success of encoder-decoder structure for various computer vision tasks~\cite{liu2017video,su2017deep,tao2017spmc,xu2017deep}, we explore the effective way to adapt it for the task of image deblurring. In this paper, we will show that directly applying an existing encoder-decoder structure cannot produce optimal results. Our Encoder-decoder ResBlock network, on the contrary, amplifies the merit of various CNN structures and yields the feasibility for training. It also produces a very large receptive field, which is of vital importance for large-motion deblurring.

Our experiments show that with the recurrent structure and combining above advantages, our end-to-end deep image deblurring framework can greatly improve training efficiency ($\approx$1/4 training time of \cite{nah2017deep} to accomplish similar restoration). We only use less than 1/3 trainable parameters with much faster testing time. Besides training efficiency, our method can also produce higher quality results than existing methods both quantitatively and qualitatively, as shown in Fig.~\ref{fig:teaser} and to be elaborated later. We name this framework {\it scale-recurrent network} (SRN).

\begin{figure*}[ht]
    \begin{center}
      \includegraphics[width=1.0\linewidth]{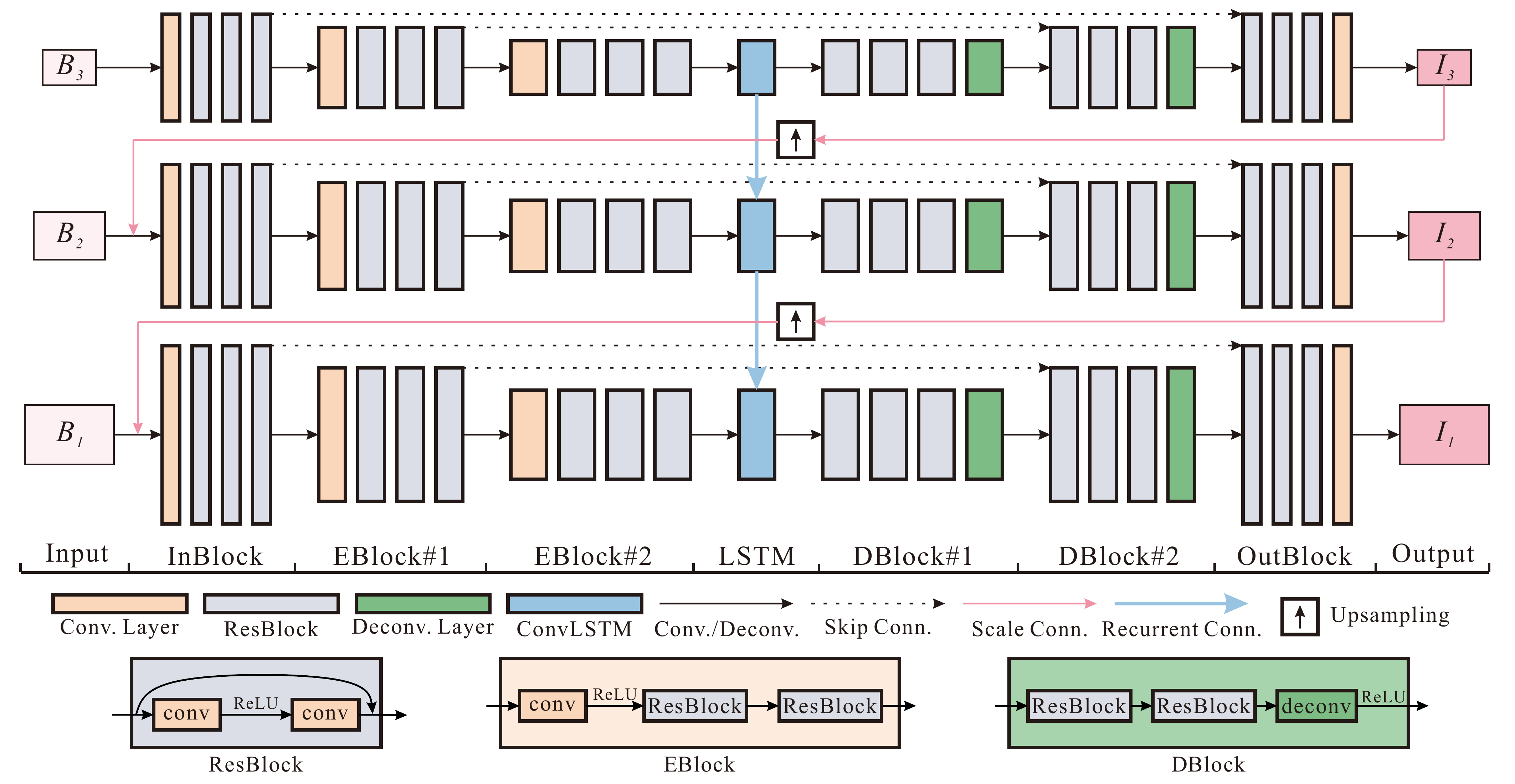}
    \end{center}
    \vspace{-0.1in}
    \caption{Our proposed SRN-DeblurNet framework.}\label{fig:framework}
    \vspace{-0.2in}
\end{figure*}

\section{Related Work} \label{sec:related}
In this section, we briefly review image deblurring methods and recent CNN structures for image processing.

\vspace{-0.1in}
\paragraph{Image/Video Deblurring}
After the seminal work of Fergus \etal~\cite{fergus2006removing} and Shan \etal~\cite{shan2008high}, many deblurring methods were proposed towards both restoration quality and adaptiveness to different situations. Natural image priors were designed to suppress artifacts and improve quality. They include total variation (TV) \cite{chan1998total}, sparse image priors \cite{levin2009understanding}, heavy-tailed gradient prior \cite{shan2008high}, hyper-Laplacian prior \cite{krishnan2009fast}, $l_0$-norm gradient prior \cite{xu2013unnatural}, \etc~
Most of these traditional methods follow the coarse-to-fine framework. Exceptions include frequency-domain methods \cite{delbracio2015burst,goldstein2012blur}, which are only applicable to limited situations.

Image deblurring also benefits from recent advance of deep CNN. Sun \etal~\cite{sun2015learning} used the network to predict blur direction. Schuler \etal~\cite{schuler2016learning} stacked multiple CNNs in a coarse-to-fine manner to simulate iterative optimization. Chakrabarti \cite{chakrabarti2016neural} predicted deconvolution kernel in frequency domain. These methods follow the traditional framework with several parts replaced to the CNN version. Su \etal~\cite{su2017deep} used an encoder-decoder network with skip-connections to learn video deblurring. Nah \etal~\cite{nah2017deep} trained a multi-scale deep network to progressively restore sharp images. These end-to-end methods make use of multi-scale information via different structures.

\vspace{-0.1in}
\paragraph{CNNs for Image Processing}
Different from classification tasks, networks for image processing require special design. As one of the earliest methods, SRCNN \cite{dong2014learning} used 3 flat convolution layers (with the same feature map size) for super-resolution. Improvement was yielded by U-net \cite{ronneberger2015u} (as shown in Fig.~\ref{fig:cnns}(a)), also termed as encoder-decoder networks \cite{mao2016image}, which greatly increases regression ability and is widely used in recent work of FlowNet \cite{dosovitskiy2015flownet}, video deblurring \cite{su2017deep}, video super-resolution \cite{tao2017spmc}, frame synthesis \cite{liu2017video}, \etc~Multi-scale CNN \cite{nah2017deep} and cascaded refinement network (CRN) \cite{chen2017photographic} (Fig.~\ref{fig:cnns}(b)) simplified training by progressively refining output starting from a very small scale. They are successful in image deblurring and image synthesis, respectively. Fig.~\ref{fig:cnns}(c) shows a different structure \cite{chen2017fast} that used dilated convolution layers with increasing rates, which approximates increasing kernel sizes.

\section{Network Architecture}\label{sec:network}

The overall architecture of the proposed network, which we call SRN-DeblurNet, is illustrated in Fig.~\ref{fig:framework}. It takes as input a sequence of blurry images downsampled from the input image at different scales, and produces a set of corresponding sharp images. The sharp one at the full resolution is the final output.

\subsection{Scale-recurrent Network (SRN)}
As explained in Sec.~\ref{sec:intro}, we adopt a novel recurrent structure across multiple scales in the coarse-to-fine strategy. We form the generation of a sharp latent image at each scale as a sub-problem of the image deblurring task, which takes a blurred image and an initial deblurred result (upsampled from the previous scale) as input, and estimates the sharp image at this scale as
\begin{equation}
   \begin{aligned}
      \mathbf{I}^i, \mathbf{h}^i &= \mathbf{Net}_{SR}(\mathbf{B}^i, \mathbf{I}^{i+1\uparrow},\mathbf{h}^{i+1\uparrow}; \theta_{SR}),\\
   \end{aligned}\label{eq:scale_recurrent}
\end{equation}
where $i$ is the scale index, with $i=1$ representing the finest scale. $\mathbf{B}^i$, $\mathbf{I}^i$ are the blurry and estimated latent images at the $i$-th scale,
respectively. $\mathbf{Net}_{SR}$ is our proposed scale-recurrent network with training parameters denoted as $\theta_{SR}$. Since the network is recurrent, hidden state features $\mathbf{h}^i$ flow across scales. The hidden state captures image structures and kernel information from the previous coarser scales. $(\cdot)^\uparrow$ is the operator to adapt features or images from the $(i-1)$-th to $i$-th scale.

Eq.~\eqref{eq:scale_recurrent} gives a detailed definition of the network. In practice, there is enormous flexibility in network design. First, recurrent networks can take different forms, such as vanilla RNN, long-short term memory (LSTM) \cite{hochreiter1997long,xingjian2015convolutional} and gated recurrent unit (GRU) \cite{chung2014empirical}. We choose ConvLSTM \cite{xingjian2015convolutional} since we found it performs better in our experiments that will be described more in Sec.~\ref{sec:exp}. Second, possible choices for operator $(\cdot)^\uparrow$ include deconvolution layer, sub-pixel convolutional \cite{shi2016real} and image resizing. We use bilinear interpolation for all our experiments for its sufficiency and simplicity. Third, the network at each scale needs to be properly designed for optimal effectiveness at recovering sharp images. Our method will be detailed in the following sections.


\subsection{Encoder-decoder with ResBlocks}
\paragraph{Encoder-decoder Network} Encoder-decoder network \cite{mao2016image,ronneberger2015u} refers to those symmetric CNN structures that first progressively transform input data into feature maps with smaller spatial size and more channels (encoder), and then transform them back to the shape of the input (decoder). Skip-connections between corresponding feature maps in encoder-decoder are widely used to combine different levels of information. They can also benefit gradient propagation and accelerate convergence. Typically, the encoder module contains several stages of convolution layers with strides, and the decoder module is implemented using a series of deconvolution layers \cite{liu2017video,su2017deep,tao2017spmc} or resizing. Additional convolution layers are inserted after each level to further increase depth.

The encoder-decoder structure has been proven to be effective in many vision tasks \cite{liu2017video,su2017deep,tao2017spmc,xu2017deep}. However, directly using the encoder-decoder network is not the best choice for our task with the following considerations. 

First, for the task of deblurring, the receptive field needs to be large to handle severe motion, resulting in stacking more levels for encoder/decoder modules. However, this strategy is not recommended in practice since it increase the number of parameters quickly with the large number of intermediate feature channels. Besides, the spatial size of middle feature map would be too small to keep spatial information for reconstruction. Second, adding more convolution layers at each level of encoder/decoder modules would make the network slow to converge (with flat convolution at each level). Finally, our proposed structure requires recurrent modules with hidden states inside.

\vspace{-0.1in}
\paragraph{Encoder/decoder ResBlock}
We make several modifications to adapt encoder-decoder networks into our framework. First, we improve encoder/decoder modules by introducing residual learning blocks \cite{he2016deep}. According to results of \cite{nah2017deep} and also our extensive experiments, we choose to use ResBlocks instead of the original building block in ResNet \cite{he2016deep} (without batch normalization). As illustrated in Fig.~\ref{fig:framework}, our proposed Encoder ResBlocks (EBlocks) contains one convolution layer followed by several ResBlocks. The stride for convolution layer is 2. It doubles the number of kernels of the previous layer and downsamples the feature maps to half size. Each of the following ResBlocks contains 2 convolution layers. Besides, all convolution layers have the same number of kernels. Decoder ResBlock (DBlocks) is symmetric to EBlock. It contains several ResBlocks followed by 1 deconvolution layer. The deconvolution layer is used to double the spatial size of features maps and halve channels.

Second, our scale-recurrent structure requires recurrent modules inside networks. Similar to the strategy of \cite{tao2017spmc}, we insert convolution layers in the bottleneck layer for hidden state to connect consecutive scales. Finally, we use large convolution kernels of size $5\times5$ for every convolution layer. The modified network is expressed as
\begin{equation}
   \begin{aligned}
      \mathbf{f}^i &= \mathbf{Net}_{E}(\mathbf{B}^i,\mathbf{I}^{i+1\uparrow}), \\
      \mathbf{h}^i,\mathbf{g}^i &= \mathbf{ConvLSTM}(\mathbf{h}^{i+1\uparrow},\mathbf{f}^i; \theta_{LSTM}),\\
      \mathbf{I}^i &= \mathbf{Net}_{D}(\mathbf{g}^i; \theta_D),
   \end{aligned}\label{eq:edresblock}
\end{equation}
where $\mathbf{Net}_{E}$ and $\mathbf{Net}_{D}$ are encoder and decoder CNNs with parameters $\theta_E$ and $\theta_D$. 3 stages of EBlocks and DBlocks are used in $\mathbf{Net}_{E}$ and $\mathbf{Net}_{D}$, respectively. $\theta_{LSTM}$ is the set of parameters in ConvLSTM. Hidden state $h^i$ may contain useful information about intermediate result and blur patterns, which is passed to the next scale and benefits the fine-scale problem.

The details of model parameters are specified here. Our SRN contains 3 scales. The $(i+1)$-th scale is half of the size of the $i$-th scale. For the encoder/decoder ResBlock network, there are 1 InBlock, 2 EBlocks, followed by 1 Convolutional LSTM block, 2 DBlocks and 1 OutBlock, as shown in Fig.~\ref{fig:framework}. InBlock produces 32-channel feature map. And OutBlock take previous feature map as input and generate output image. The numbers of kernels of all convolution layers inside each EBlock/DBlock are the same. For EBlocks, the numbers of kernels are 64 and 128, respectively. For DBlocks, they are 128 and 64. The stride size for the convolution layer in EBlocks and deconvolution layers is 2, while all others are 1. Rectified Linear Units (ReLU) are used as the activation function for all layers, and all kernel sizes are set to 5.

\subsection{Losses}
We use Euclidean loss for each scale, between network output and the ground truth (downsampled to the same size using bilinear interpolation) as
\begin{align}
   \mathcal{L}=\sum_{i=1}^n \frac{\kappa_i}{N_i}\Vert I^i - I^i_*\Vert^2_2\label{eq:loss},
\end{align}
where $I^{i}$ and $I^i_*$ are our network output and ground truth respectively in the $i$-th scale. $\{\kappa_i\}$ are the weights for each scale. We empirically set $\kappa_{i}=1.0$. $N_i$ is the number of elements in $I^i$ to normalize. We have also tried total variation and adversarial loss. But we notice that $L_2$-norm is good enough to generate sharp and clear results.

\section{Experiments}\label{sec:exp}
Our experiments are conducted on a PC with Intel Xeon E5 CPU and an NVIDIA Titan X GPU. We implement our framework on TensorFlow platform \cite{tensorflow2015-whitepaper}. Our evaluation is comprehensive to verify different network structures, as well as various network parameters. For fairness, unless noted otherwise, all experiments are conducted on the same dataset with the same training configuration.

\vspace{-0.05in}
\paragraph{Data Preparation}
To create a large training dataset,
early learning-based methods \cite{chakrabarti2016neural,schuler2016learning,sun2015learning} synthesize blurred images by convolving sharp images with real or generated uniform/non-uniform blur kernels. Due to the simplified image formation models, the synthetic data is still different from real ones that are captured by cameras. Recently, researchers \cite{nah2017deep,su2017deep} proposed generating blurred images through averaging consecutive short-exposure frames from videos captured by high-speed cameras, e.g. GoPro Hero 4 Black, to approximate long-exposure blurry frames. These generated frames are more realistic since they can simulate complex camera shake and object motion, which are common in real photographs.

For fair comparison with respect to the network structure, we train our network using the GOPRO dataset proposed in \cite{nah2017deep}, which contains 3,214 blurry/clear image pairs. Following the same strategy as in \cite{nah2017deep}, we use 2,103 pairs for training and the remaining 1,111 pairs for evaluation.

\vspace{-0.05in}
\paragraph{Model Training}
For model training, we use Adam solver \cite{kingma2014adam} with $\beta_1=0.9$, $\beta_2=0.999$ and $\epsilon=10^{-8}$. The learning rate is exponentially decayed from initial value of $0.0001$ to $1e^{-6}$ at $2000$ epochs using power $0.3$. According to our experiments, 2,000 epochs are enough for convergence, which takes about 72 hours. In each iteration, we sample a batch of 16 blurry images and randomly crop $256\times256$-pixel patches as training input. Ground truth sharp patches are generated accordingly. All trainable variables are initialized using Xavier method \cite{glorot2010understanding}. The parameters described above are fixed for all experiments. 

For experiments that involve recurrent modules, we apply gradient clip only to weights of ConvLSTM module (clipped by global norm 3) to stabilize training. Since our network is fully convolutional, images of arbitrary size can be fed in it as input, as long as GPU memory allows. For testing image of size $720\times1280$, running time of our proposed method is around 1.6 seconds.

\begin{figure}[h]
  \begin{center}
   \includegraphics[width=1.0\linewidth]{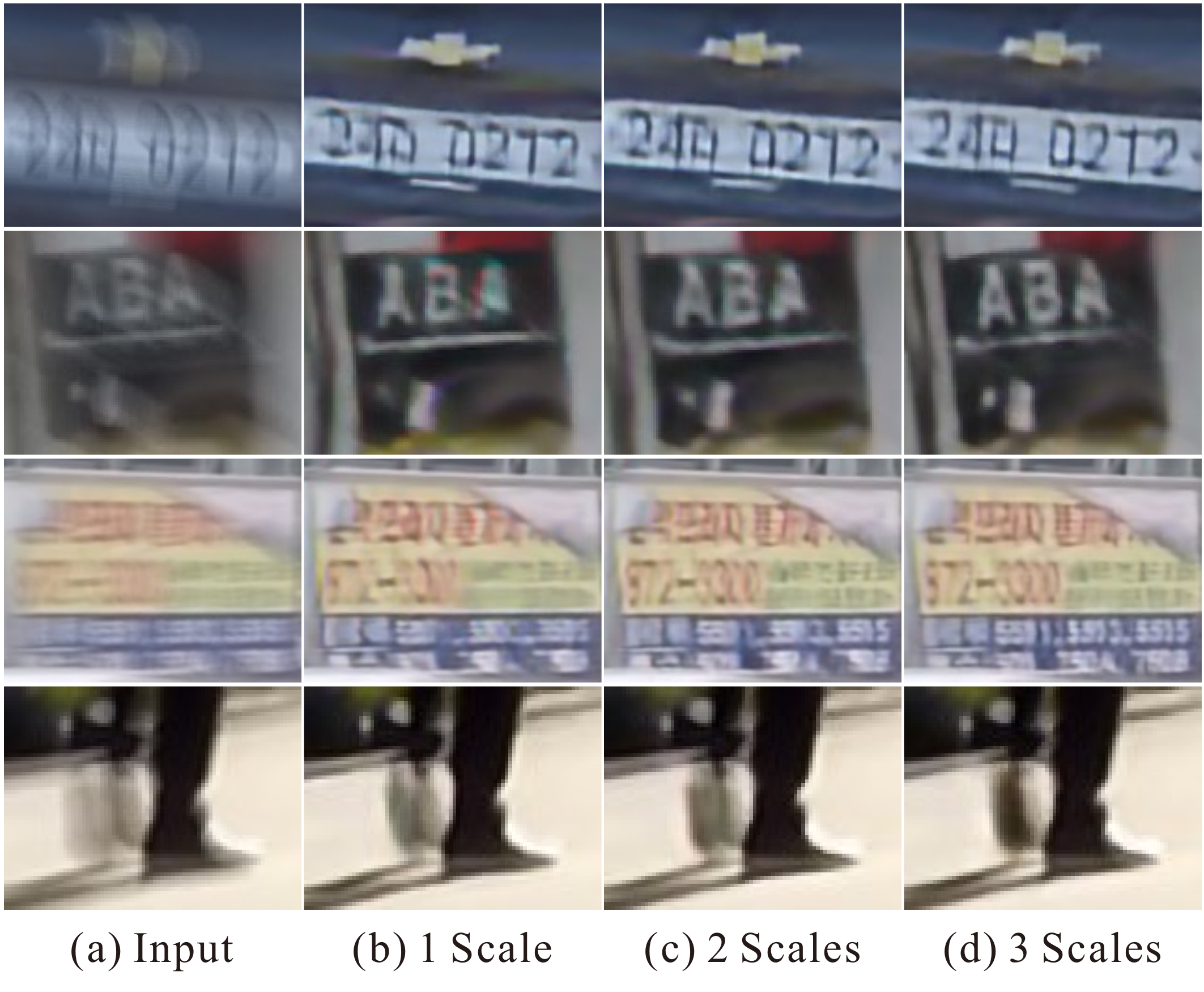}
  \end{center}
  \vspace{-0.1in}
  \caption{Results of multi-scale baseline method.}\label{fig:exp_multiscale}
  \vspace{-0.2in}
\end{figure}

\begin{table}[h]
    \center
    \caption{Quantitative results for baseline models.}\label{tab:exp}
    \vspace{0.1in}
    \setlength{\tabcolsep}{4pt}
    \footnotesize
    \begin{tabular}{c||ccccc}
       \hline
       Model & SS    & SC    & w/o R & RNN   & SR-Flat  \\ \hline
       Param & 2.73M & 8.19M & 2.73M & 3.03M & 2.66M    \\ \hline
       PSNR  & 28.40 & 29.05 & 29.26 & 29.35 & 27.53    \\ \hline
       SSIM  & 0.9045& 0.9166& 0.9197& 0.9210& 0.8886   \\ \hline
       \hline
       Model & SR-RB  & SR-ED   & SR-EDRB1 & SR-EDRB2 & SR-EDRB3\\ \hline
       Param & 2.66M  & 3.76M   & 2.21M    & 2.99M    & 3.76M   \\ \hline
       PSNR  & 28.11  & 29.06   & 28.60    & 29.32    & \textbf{29.98} \\ \hline
       SSIM  & 0.8991 & 0.9170  & 0.9082   & 0.9204   & \textbf{0.9254} \\ \hline
    \end{tabular}
    \vspace{-0.1in}
\end{table}

\subsection{Multi-scale Strategy}
To evaluate the effectiveness of the proposed scale-recurrent network, we design several baseline models. Note to evaluate network structures, we use kernel size 3 for all convolution layers, for the efficiency's sake. Single-scale model \textbf{SS} uses the same structure as our proposed one, except that only a single-scale image is taken as input at its original resolution. Recurrent modules are replaced by one convolution layer to ensure the same number of convolution layers. 

Baseline model \textbf{SC} refers to the scale-cascaded structure as in \cite{chen2017photographic,nah2017deep}, which uses 3 stages of independent networks. Each single-stage network is the same as model \textbf{SS}. Therefore, the trainable parameters of this model are 3 times more than our method. Model \textbf{w/oR} does not contain explicit recurrent modules in bottleneck layer (\ie~model \textbf{SS}), which is a shared-weight version of model \textbf{SC}. Model \textbf{RNN} uses vanilla RNN structure instead of ConvLSTM.

The results of different methods on the testing dataset are shown in Table~\ref{tab:exp}, from which we make several useful observations. First, the multi-scale strategy is very effective for the image deblurring task. Model \textbf{SS} uses the same structure and the same number of parameters as our proposed SRN structure, and yet performs much worse in terms of PSNR (28.40dB \vs 29.98dB). 
One visual comparison is given in Fig.~\ref{fig:exp_multiscale} where the single-scale Model \textbf{SS} in (b) can recover structure from severely blurred input. But the characters are still not clear enough for recognition. 

Results are improved when we use 2 scales as shown in Fig.~\ref{fig:exp_multiscale}(c), because multi-scale information has been effectively incorporated. The more complete model with 3 scales further produces better results in Fig.~\ref{fig:exp_multiscale}(d); but the improvements are already minor.

Second, independent parameters for each scale are \emph{not} necessary and may be even harmful, proved by the fact that Model \textbf{SC} performs worse than Model \textbf{w/oR}, \textbf{RNN} and \textbf{SR-EDRB3} (which share the same Encoder-decoder ResBlock structure with 3 ResBlocks). We believe the reason is that, although more parameters lead to a larger model capacity, it also requires longer training time and larger training dataset. In our constrained settings of fixed dataset and training epochs, Model \textbf{SC} may not be optimally trained. 

Finally, we also test different recurrent modules. The results show that vanilla RNN is better than not using RNN, and ConvLSTM achieves the best results with model \textbf{SR-EDRB3}.

\subsection{Encoder-decoder ResBlock Network}
We also design a series of baseline models to evaluate the effectiveness of the encoder-decoder with ResBlock structure. For fair comparison, all models here use our scale-recurrent (\textbf{SR}) framework. Model \textbf{SR-Flat} replaces encoder-decoder architecture with flat convolution layers, the number of which is the same as the proposed network, \ie~43 layers. Model \textbf{SR-RB} replaces all EBlocks and DBlocks with ResBlock. No stride or pooling is included. This makes feature maps have the same size. Model \textbf{SR-ED} uses original encoder-decoder structure, with all ResBlocks replaced by 2 convolution layers. We also compare with different numbers of ResBlocks in EBlock/DBlock. Models \textbf{SR-EDRB1}, \textbf{SR-EDRB2} and \textbf{SR-EDRB3} refer to 1, 2 and 3 ResBlocks models, respectively.

Quantitative results are shown in Table~\ref{tab:exp}. Flat convolution model \textbf{Flat} performs worst in terms of both PSNR and SSIM. In our experiments, it takes significantly more time to reach the same level of quality as other results. Model \textbf{RB} is much better, since ResBlock structure is designed for better training. The best results are accomplished by our proposed model \textbf{SR-EDRB1-3}. The quantitative results also get better as the number of ResBlocks increases. We choose 3 ResBlocks in our proposed model, since the improvement beyond 3 ResBlocks is marginal and it is a good balance between efficiency and performance.

\subsection{Comparisons}
We compare our method with previous state-of-the-art image deblurring approaches on both evaluation datasets and real images. Since our model deals with general camera shake and object motion (\ie~dynamic deblurring \cite{hyun2013dynamic}), it is unfair to compare with traditional uniform deblurring methods. The method of Whyte \etal~\cite{whyte2012non} is selected as a representative traditional method for non-uniform blur. Note that for most examples in the testing dataset, blurred images are caused merely by camera shake. Thus the non-uniform assumption in \cite{whyte2012non} holds. 

\begin{figure*}[h]
    \begin{center}
      \includegraphics[width=1.0\linewidth]{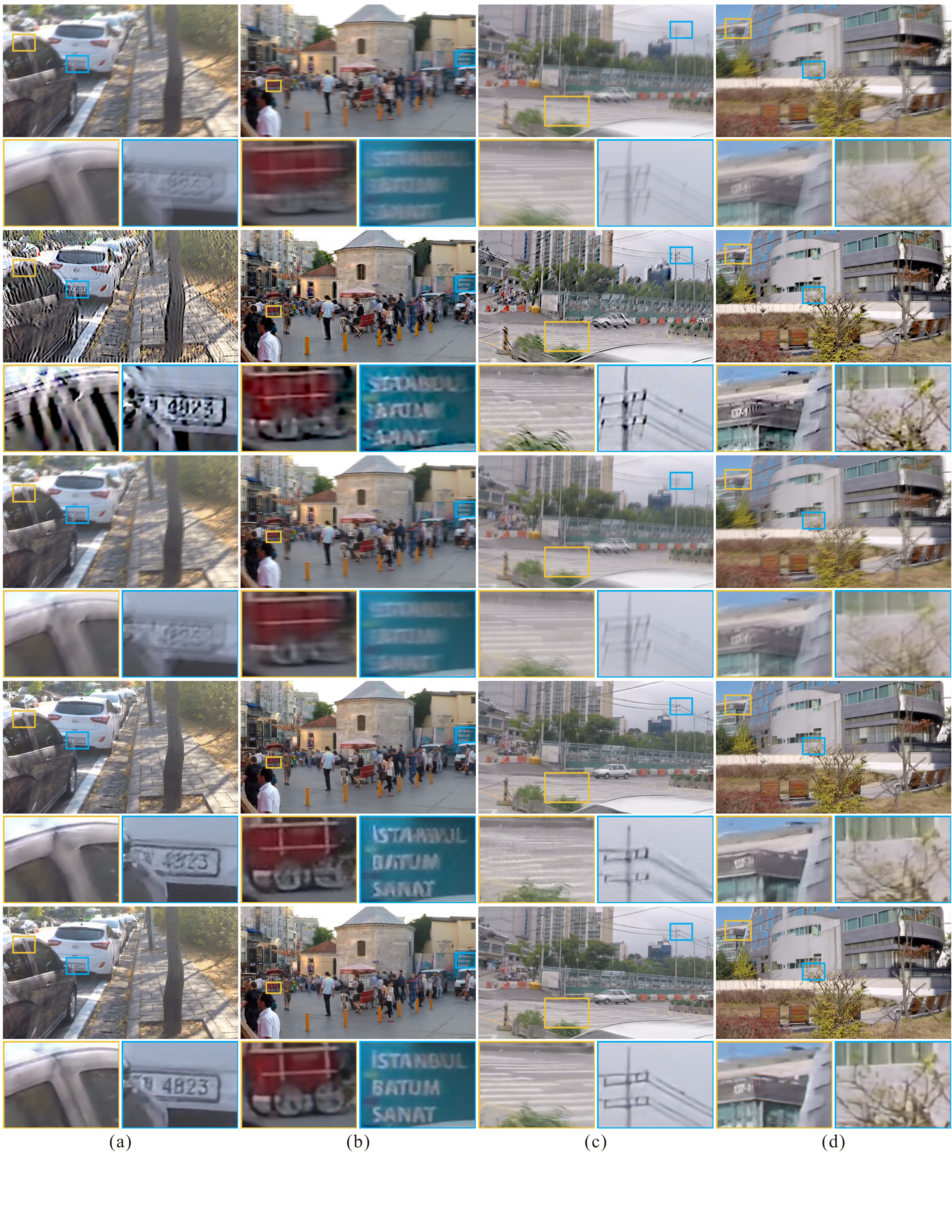}
    \end{center}
   \caption{\textbf{Visual comparisons on testing dataset.} From \textbf{Top} to \textbf{Bottom}: input, Whyte \etal~\cite{whyte2012non}, Sun \etal~\cite{sun2015learning}, Nah \etal~\cite{nah2017deep} and ours (best to zoom-in and view on screen).}\label{fig:comp_deep}
\end{figure*}

\begin{figure*}[h]
  \begin{center}
    \includegraphics[width=1.0\linewidth]{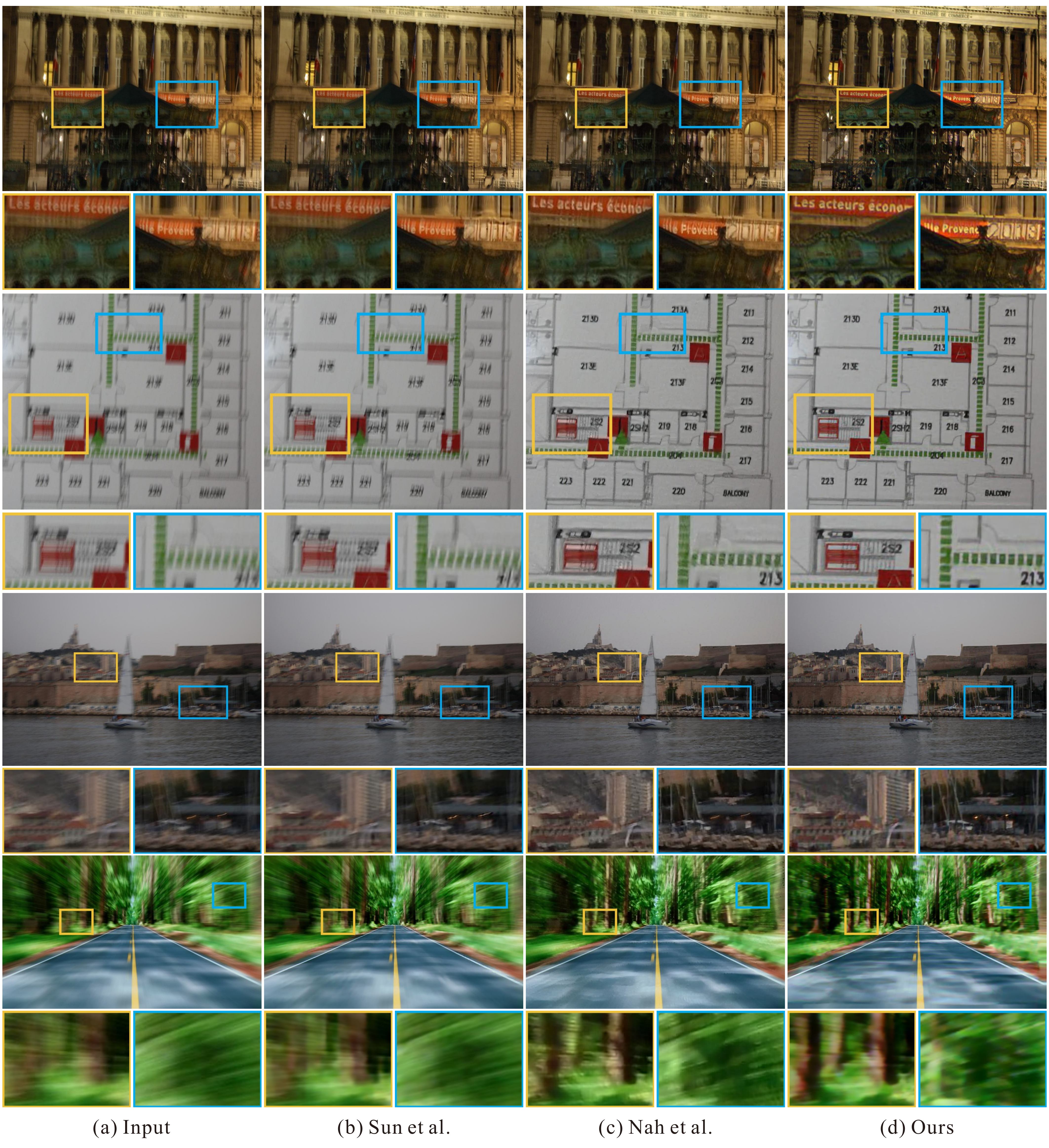}
  \end{center}
  \vspace{-0.17in}
  \caption{Real blurred images.}\label{fig:real}
\end{figure*}

Kim \etal~\cite{hyun2013dynamic} can handle complex dynamic blurring, but does not provide code or results. Instead we compare with more recent work of Nah \etal~\cite{nah2017deep}, which demonstrated very good results. Sun \etal~\cite{sun2015learning} estimated blur kernels using CNN, and applied traditional deconvolution methods to recover the sharp image. We use official implementation from the authors with default parameters. The quantitative results on GOPRO testing set and K{\"o}hler Dataset \cite{kohler2012recording} are listed in Table~\ref{tab:comp}. Visual comparison is shown in Figs.~\ref{fig:comp_deep} and \ref{fig:real}. More results are included in our supplementary material. 

\begin{table}
    \center
    \caption{Quantitative results on testing dataset (PSNR/SSIM).}\label{tab:comp}
    \vspace{0.1in}
    \footnotesize
    \begin{tabular}{c|c|c|c|c|c}
       \hline
       \multirow{2}{*}{Method}&\multicolumn{2}{c|}{GOPRO} &  \multicolumn{2}{c|}{K{\"o}hler Dataset}&\multirow{2}{*}{Time}\\\cline{2-5}
                   & PSNR  & SSIM   & PSNR  & MSSIM          \\\hline
       Kim \etal   & 23.64 & 0.8239 & 24.68 & 0.7937 & 1 hr \\\hline
       Sun \etal   & 24.64 & 0.8429 & 25.22 & 0.7735 & 20 min\\\hline
       Nah \etal   & 29.08 & 0.9135 & 26.48 & 0.8079 & 3.09 s\\\hline
       Ours        & \textbf{30.10} & \textbf{0.9323} & \textbf{26.80} & \textbf{0.8375} & \textbf{1.6s}\\\hline
    \end{tabular}
    \vspace{-0.1in}
\end{table}

\paragraph{Benchmark Datasets} The first row of Fig.~\ref{fig:comp_deep} contains images from the testing datasets, which suffer from complex blur due to large camera and object motion. Although traditional method  \cite{whyte2012non} models a general non-uniform blur for camera translation and rotation, it still fails for Fig.~\ref{fig:comp_deep}(a), (c), and (d), where camera motion dominates. It is because forward/backward motion, as well as scene depth, plays important roles in real blurred images. Moreover, violation of the assumed model results in annoying ringing artifacts, which make restored image even worse than input. 

Sun \etal~used CNN to predict kernel direction. But on this dataset, the complex blur patterns are quite different from their synthetic training set. Thus this method failed to predict reliable kernels on most cases, and results are only slightly sharpened. Recent state-of-the-art method \cite{nah2017deep} can produce good quality results, with remaining a few blurry structure and artifacts. Thanks to the designed framework and modules, our method produces superior results with sharper structures and clear details. According to our experiments, even on extreme cases, where motion is too large for previous solutions, our method can still produce reasonable results for important part and does not cause much artifacts on other regions, as shown in the last case of Fig.~\ref{fig:real}. Quantitative results also validate our observation, where our framework outperforms others by a large margin.

\paragraph{Real Blurred Images} Previous testing images are synthesized from high-speed cameras, which may still differ from real blurred inputs. We show our results on real-captured blurred images in Fig.~\ref{fig:real}. Our trained model generalizes well on these images, as shown in Fig.~\ref{fig:real}(d). Compared with Sun \etal and Nah \etal, our results are of high quality.

\section{Conclusion}
\vspace{-0.09in}
In this paper, we have explained what is the proper network structure for using the ``coarse-to-fine'' scheme in image deblurring. 
We have also proposed a scale-recurrent network, as well as an encoder-decoder ResBlocks structure in each scale.
This new network structure has less parameters than previous multi-scale deblurring ones and is easier to train. The results generated by our method are state-of-the-arts, both qualitatively and quantitatively. We believe this scale-recurrent network can be applied to other image processing tasks, and we will explore them in the future.

{\small
\bibliographystyle{ieee}
\bibliography{deblur}
}

\end{document}